\documentclass{article}

\usepackage{PRIMEarxiv}

\usepackage[utf8]{inputenc} 
\usepackage[T1]{fontenc}    
\usepackage{hyperref}       
\usepackage{url}            
\usepackage{booktabs}       
\usepackage{amsfonts}       
\usepackage{nicefrac}       
\usepackage{microtype}      
\usepackage{lipsum}
\usepackage{fancyhdr}       
\usepackage{graphicx}       
\graphicspath{{media/}}     

\title{Efficient Implementation of the Global Cardinality Constraint with Costs
}

\author{
  Margaux SCHMIED \\
  Université Côte d'Azur, CNRS, I3S, France \\
  \texttt{margaux.schmied@icloud.com} \\
   \And
  Jean-Charles REGIN \\
  Université Côte d'Azur, CNRS, I3S, France \\
  \texttt{jcregin@gmail.com} \\
}

\usepackage{macro}

\begin{document}
\maketitle

\begin{abstract}
The success of Constraint Programming relies partly on the global constraints and implementation of the associated filtering algorithms. Recently, new ideas emerged to improve these implementations in practice, especially regarding the all different constraint.

In this paper, we consider the cardinality constraint with costs. The cardinality constraint is a generalization of the all different constraint that specifies the number of times each value must be taken by a given set of variables in a solution. The version with costs introduces an assignment cost and bounds the total sum of assignment costs. The arc consistency filtering algorithm of this constraint is difficult to use in practice, as it systematically searches for many shortest paths. We propose a new approach that works with upper bounds on shortest paths based on landmarks. This approach can be seen as a preprocessing. It is fast and avoids, in practice, a large number of explicit computations of shortest paths.
\end{abstract}


\section{Introduction} 

In Constraint Programming (CP), a problem is defined on variables and constraints. Each variable is provided with a domain defining the set of its possible values. A constraint expresses a property that must be satisfied by a set of variables. CP uses a specific resolution method for each constraint. 

The success of CP relies on the use of high-performance filtering algorithms (also known as propagators). These algorithms remove values from variable domains that are not consistent with the constraint, i.e. that do not belong to a solution of the constraint's underlying sub-problem.
The most well-known propagator is that of the all different (alldiff) constraint, which specifies that a set of variables must all take different values. The efficiency in practice of that propagator strongly depends on its implementation. Thus, algorithms proposing practical improvements on Régin's algorithm~\cite{Regin:AfilteringalgorithmforconstraintsofdifferenceinCSPs} are still appearing~\cite{Zhang:AFastAlgorithmforGeneralizedArcConsistencyoftheAlldifferentConstraint,Zhang:EarlyandEfficientIdentificationofUselessConstraintPropagationforAlldifferentConstraints}.

In this article, we consider another constraint introduced by Régin that is also popular~\cite{Demassey:ACost-RegularBasedHybridColumnGenerationApproach,Nightingale:AutomaticallyimprovingconstraintmodelsinSavileRow,vanHoeve:Onglobalwarming:Flow-basedsoftglobalconstraints,Gualandi:ConstraintProgramming-basedColumnGeneration,Ducomman:AlternativeFilteringfortheWeightedCircuitConstraint:ComparingLowerBoundsfortheTSPandSolvingTSPTW}: the cardinality constraint with costs~\cite{Regin:CostbasedArcConsistencyforGlobalCardinalityConstraints}. 
We propose to try to speed up its filtering algorithm when there is nothing to deduce.
This is often the case at the start of the search, particularly as the optimal value is far from known. In addition, at this stage, the gains can be significant since few values have been removed from the domains, and so the complexity of the algorithms is greater. This approach can be particularly interesting with aggressive restarting methods and could simplify the use of CP: there is less need to worry about the inference strength of constraints versus their cost. We can worry less about the type of filtering to be used and consider the arc consistency right away.

The global cardinality constraint (gcc)~\cite{Regin:gcc} is a generalization of the alldiff constraint. 
A gcc is specified in terms of a set of variables $X=\{x_1,...,x_p\}$
which take their values in a subset of $V=\{v_1,...,v_d\}$. It constrains the number of times a value $v_i \in V$ is assigned to variables in $X$ to belong to an interval $[l_i,u_i]$.

A gcc with costs (costgcc) is a generalization of a gcc in which a
cost is associated with each value of each variable. Then, each solution of the underlying gcc is associated with a 
global cost equal to the sum of the costs associated with the assigned values of the
solution. In a costgcc constraint the global cost must be less than a given value, H.

\begin{figure}[h!]
    \centering
    \begin{subfigure}[t]{0.4\columnwidth}
        \begin{tikzpicture}[
        round/.style={regular polygon, regular polygon sides=6, draw=black},
        square/.style={rectangle, draw=black},
        oval/.style={rounded rectangle, draw=black, minimum width=1.5cm, minimum height=0.7cm},
        flecheG/.style={stealth-},
        flecheD/.style={-stealth},
        align=center
        every node/.style={transform shape}
        ]
    
    \node[oval] (x1) {Peter};
    \node[oval] (x2)            [below of=x1]         {Paul};
    \node[oval] (x3)            [below of=x2]         {Mary};
    \node[oval] (x4)            [below of=x3]         {John};
    \node[oval] (x5)            [below of=x4]         {Bob};
    \node[oval] (x6)            [below of=x5]         {Mike};
    \node[oval] (x7)            [below of=x6]         {Julia};

    \node (w1) [right of=x1, yshift=-0.1cm, xshift= -0.2cm] {1};
    \node (w2) [right of=x1, yshift=-0.55cm, xshift= -0.1cm] {4};
    \node (w3) [right of=x2, yshift=0.2cm, xshift= -0.2cm] {1};
    \node (w4) [right of=x2, yshift=-0.5cm, xshift= -0.2cm] {4};
    \node (w5) [right of=x3, yshift=0.17cm, xshift= -0.1cm] {3};
    \node (w6) [right of=x3, yshift=-0.15cm, xshift= -0.2cm] {1};
    \node (w7) [right of=x4, yshift=0.54cm, xshift= -0.1cm] {3};
    \node (w8) [right of=x4, yshift=0.1cm, xshift= -0.2cm] {1};
    \node (w9) [right of=x5, yshift=0.5cm, xshift= -0.2cm] {1};
    \node (w10) [right of=x6, yshift=0.5cm, xshift= -0.2cm] {1};
    \node (w11) [right of=x7, yshift=0.9cm, xshift= -0.2cm] {1};
    \node (w12) [right of=x7, yshift=-0cm, xshift= -0.2cm] {1};

    \node[square] (x8) [right of=x1, yshift=-1cm, xshift= 1.5cm] {A};
    \node[square] (x9) [below of=x8] {B};
    \node[square] (x10) [below of=x9] {C};
    \node[square] (x11) [below of=x10] {D};
    \node[square] (x12) [below of=x11] {E};

    \node (c1) [right of=x8, xshift= -0.2cm] {[1, 2]};
    \node (c2) [right of=x9, xshift= -0.2cm] {[1, 2]};
    \node (c3) [right of=x10, xshift= -0.2cm] {[1, 1]};
    \node (c4) [right of=x11, xshift= -0.2cm] {[0, 2]};
    \node (c5) [right of=x12, xshift= -0.2cm] {[0, 2]};
    

    \draw[-] (x8) -- (x1);
    \draw[-] (x8) -- (x2);
    \draw[-] (x8) -- (x3);
    \draw[-] (x8) -- (x4);

    \draw[-] (x9) -- (x1);
    \draw[-] (x9) -- (x2);
    \draw[-] (x9) -- (x3);
    \draw[-] (x9) -- (x4);

    \draw[-] (x10) -- (x5);

    \draw[-] (x11) -- (x6);
    \draw[-] (x11) -- (x7);

    \draw[-] (x12) -- (x7);

    \end{tikzpicture} 
    \end{subfigure}
    \hfill
    \begin{subfigure}[t]{0.4\columnwidth}
        \begin{tikzpicture}[
        round/.style={regular polygon, regular polygon sides=6, draw=black},
        square/.style={rectangle, draw=black},
        oval/.style={rounded rectangle, draw=black, minimum width=1.5cm, minimum height=0.7cm},
        flecheG/.style={stealth-},
        flecheD/.style={-stealth},
        align=center
        every node/.style={transform shape}
        ]
            
            \node[oval] (x1) {Peter};
            \node[oval] (x2) [below of=x1] {Paul};
            \node[oval] (x3) [below of=x2] {Mary};
            \node[oval] (x4) [below of=x3] {John};
            \node[oval] (x5) [below of=x4] {Bob};
            \node[oval] (x6) [below of=x5] {Mike};
            \node[oval] (x7) [below of=x6] {Julia};
    
            \node (w1) [right of=x1, yshift=-0.2cm, xshift= -0.2cm] {1};
            \node (w3) [right of=x2, yshift=0.2cm, xshift= -0.2cm] {1};
            \node (w6) [right of=x3, yshift=0.2cm, xshift= -0.2cm] {1};
            \node (w8) [right of=x4, yshift=0.2cm, xshift= -0.2cm] {1};
            \node (w9) [right of=x5, yshift=0.2cm, xshift= -0.2cm] {1};
            \node (w10) [right of=x6, yshift=0.6cm, xshift= -0.2cm] {1};
            \node (w11) [right of=x7, yshift=0.6cm, xshift= -0.2cm] {1};
            \node (w12) [right of=x7, yshift=0.1cm, xshift= -0.2cm] {1};
        
            \node[square] (x8) [right of=x1, yshift=-1cm, xshift= 1cm] {A};
            \node[square] (x9) [below of=x8] {B};
            \node[square] (x10) [below of=x9] {C};
            \node[square] (x11) [below of=x10] {D};
            \node[square] (x12) [below of=x11] {E};
    
            \node (c1) [right of=x8, xshift= -0.2cm] {[1, 2]};
            \node (c2) [right of=x9, xshift= -0.2cm] {[1, 2]};
            \node (c3) [right of=x10, xshift= -0.2cm] {[1, 1]};
            \node (c4) [right of=x11, xshift= -0.2cm] {[0, 2]};
            \node (c5) [right of=x12, xshift= -0.2cm] {[0, 2]};
            
        
            \draw[-] (x8) -- (x1);
            \draw[-] (x8) -- (x2);
        
            \draw[-] (x9) -- (x3);
            \draw[-] (x9) -- (x4);
        
            \draw[-] (x10) -- (x5);
        
            \draw[-] (x11) -- (x6);
            \draw[-] (x11) -- (x7);
        
            \draw[-] (x12) -- (x7);
        
        \end{tikzpicture} 
        \label{fig:introSuppr}
    \end{subfigure}

    \caption{Example of a global cardinality constraint with costs. 
    Source \protect\cite{Regin:CostbasedArcConsistencyforGlobalCardinalityConstraints}. 
    The sum of assignment costs must be less than or equal to 11. On the left, the original problem and on the right, the same problem after deleting all arcs that cannot belong to a solution.
    }

    \label{fig:intro}
\end{figure}
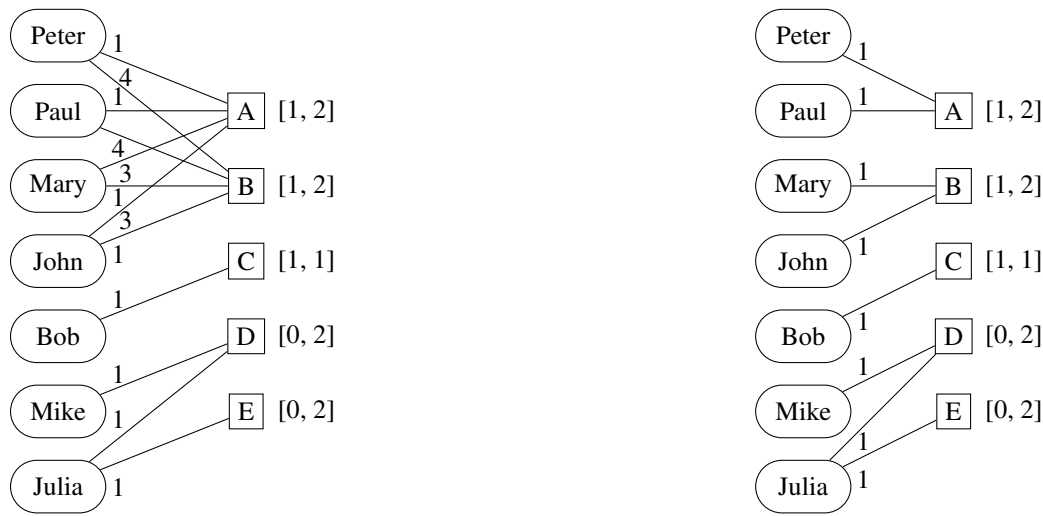

Cardinality constraints with costs has proved useful in
many real-life applications, such as routing, scheduling, 
rostering, or resource allocation problems.
The total costs are often used for expressing preferences, time or cost.

Figure~\ref{fig:intro} gives an example of a costgcc constraint and the associated filtering algorithm.
There are $7$ workers represented by the variables ${Peter, Paul, Mary, John, Bob, Mike, Julia}$ and $5$ tasks represented by the values ${A, B, C, D, E}$. Each worker has the ability to perform certain tasks and must perform exactly one of them. There is an arc from a worker to a task if the worker can perform the task, its cost corresponding to the time it takes the worker to perform the task. A task has a capacity defining the number of times it must be performed. For example, $A$ must be performed between $1$ and $2$ times. The objective is to find an assignment whose sum of costs is less than $11$. The best possible assignment has a cost of $7$, so it is a solution. On the right-hand side of Figure~\ref{fig:intro}, all arcs that cannot belong to a solution have been removed. For example, the arc $(Peter, B)$ can be deleted. If $B$ is assigned to $Peter$, then the maximum capacity of $B$ will be exceeded, so the arc $(Mary, B)$ or $(John, B)$ cannot be part of the solution. If $(John, B)$ is kept, then a value must be assigned to $Mary$, the only possibility is $(Mary, A)$ with a cost of $3$. The cost of all assignments is now $12$, which is more than $11$, so this is not a solution. Similarly, if $(Mary, B)$ is kept, then the only possibility for $John$ is $(John, A)$ with a cost of $3$ and the total cost is $12$, which is too high.

The filtering algorithm associated with a costgcc constraint~\cite{Regin:CostbasedArcConsistencyforGlobalCardinalityConstraints} can be described as repetitive. First, it computes a maximum flow at minimum cost to determine whether the constraint is consistent (i.e., admits a solution). Then, to find out whether a variable $x$ can be instantiated with a value $a$, it tries to pass a unit of flow through the arc representing the assignment of $a$ to $x$ so that the total cost of the flow is less than $H$. 
This operation involves computing min-cost flow through an arc from a given min-cost flow. This can be done by searching for a shortest path between $x$ and $a$ in the residual graph of the min-cost flow. Furthermore, it has been shown that it is possible to avoid computing a shortest path for each value of each variable and that computing one shortest path per assigned value (which is less than the number of variables) is sufficient~\cite{Regin:CostbasedArcConsistencyforGlobalCardinalityConstraints}. Unfortunately, the algorithm is repeated for each assigned value, which often proves prohibitively expensive.

In this paper, we introduce a new approach avoiding this repetitive aspect as much as possible. Our approach is based on several observations:
\begin{itemize}
    \item Finding a min-cost flow for each assignment is not necessary. Finding that there exists a flow whose cost is less than $H$ is enough. 
\item 
It is not necessary to compute any path exactly because we are only interested in their costs, not the path. Further, the exact value of the cost is not necessary either. An upper bound below a maximum cost is sufficient.
\item The use of landmarks (i.e., particular nodes) have proved their worth in speeding up computations of the shortest paths between large elements (millions of nodes)~\cite{Goldberg:ComputingtheShortestPath:ASearchMeetsGraphTheory}. 
Let $x$ and $y$ be two nodes of a graph and $p$ be another node called landmark, then we have: $d(x,p)+d(p,y) \geq d(x,y)$ where $d(i,j)$ is the shortest path distance from $i$ to $j$. Thus, by selecting one or several good landmarks $p$ we can find a good upper bound of $d(x,y)$ for each pair of nodes $x$, $y$.
\item Calls to the filtering algorithm often do not remove any value. This means that the margin (i.e., slack between $H$ and the min-cost flow value) is often large relative to the data, so using the upper bound should give good results.
\end{itemize}

On the basis of the above, we propose to introduce preprocessing in order to reduce the effective shortest path computations as proposed by Régin's algorithm.
Our approach is to search for landmarks and use them to compute upper bounds on paths to avoid unnecessary explicit shortest path computations. We consider several types of landmarks to integrate the structure of the graph, such as landmarks at the periphery (outline) of the graph or at the center. The advantage of this approach is its low cost because only two shortest paths are required per landmark. We also introduce a way to quickly detect whether a costgcc constraint is arc consistent. 

The paper is organized as follows. Section~\ref{sec:Preliminaries}
recalls some preliminaries on constraint programming, graph and flow theory. Section~\ref{sec:FilteringAlgorithm} describes Régin's algorithm because our method is based on it. 
Section~\ref{sec:UpperBounds} introduces upper bounds on shortest paths based on landmarks and, in Section~\ref{sec:ImprovedFilteringAlgorithm}, the arc consistency algorithm is accordingly adapted. Section~\ref{sec:LandmarkSelection} details some landmark selection methods. Section~\ref{sec:Experimentation} 
gives some experiments on classical problems showing that our approach dramatically reduces the number of computed shortest paths. 
 
\section{Preliminaries}
\label{sec:Preliminaries}
The following definitions, theorems and algorithms are based on the following papers and books:~\cite{Regin:CostbasedArcConsistencyforGlobalCardinalityConstraints,Berge:GrapheetHypergraphes,Lawler:CombinatorialOptimization,Tarjan:DataStructuresandNetworkAlgorithms,Ahuja:NetworkFlows}.

\paragraph*{Constraint Programming:}
A finite constraint network ${\cal N}$ is defined as a set of $n \in N$ variables $X=\{x_1, \ldots, x_n\}$, a set of current domains ${\cal D} = \{D(x_1),\ldots, D(x_n)\}$ where $D(x_i)$ is the finite set of possible values for variable $x_i$, and a set ${\cal C}$ of constraints between variables. 
We introduce the particular notation ${\cal D}_0 = \{D_0(x_1),\ldots, 
D_0(x_n)\}$ to represent the set of initial domains of ${\cal N}$ 
on which constraint definitions were stated. 
A constraint $C$ on the ordered set of variables 
$X(C)=(x_{i_1},\ldots,x_{i_r})$ is a subset $T(C)$ of the Cartesian 
product $D_0(x_{i_1}) \times \cdots \times D_0(x_{i_r})$ 
that specifies the allowed combinations of values for 
the variables $x_{i_1}, \ldots, x_{i_r}$. 
An element of $D_0(x_{i_1}) \times \cdots \times D_0(x_{i_r})$ 
is called a tuple on $X(C)$ and is denoted by $\tau$. In a tuple $\tau$, the assignment of the $i^{\text{th}}$ variable is denoted by $\tau_i$. \var$(C,i)$ represents the $i^{\text{th}}$ 
variable of $X(C)$.
A value $a$ for a variable $x$ is often denoted by $(x,a)$. 
Let $C$ be a constraint. A tuple $\tau$ on $X(C)$ is valid if $\forall (x, a) \in \tau, a \in D(x)$. $C$ is consistent iff there exists a tuple $\tau$ of $T(C)$ which is valid. A value $a \in D(x)$ is consistent with $C$ iff $x \not \in X(C)$ or there exists a valid tuple $\tau$ of $T(C)$ with $(x,a) \in \tau$. 

The costgcc constraint is formally defined as follows.

\begin{definition}[\cite{Regin:CostbasedArcConsistencyforGlobalCardinalityConstraints}]
A {\bf global cardinality constraint with costs} is a constraint
$C$ associated with a cost function on $X(C)$ \texttt{cost}, an integer $H$
and in which each value $a_i \in D(X(C))$ is associated with two
positive integers $l_i$ and $u_i$\\
\begin{tabular}{@{\hspace{0.1cm}}l@{\hspace{0.1cm}}l}
$T(C)=\{$&$\tau$ such that $\tau$ is a tuple on $X(C)$\\ 
& and $\forall a_{i} \in D(X(C)) : l_{i} \leq \#(a_{i},\tau) \leq u_{i}$\\
& and $\Sigma_{i=1}^{|X(C)|} cost($\var$(C,i),\tau[i]) \leq H$ $\}$\\
\end{tabular}\\
It is denoted by $costgcc(X,l,u,cost,H)$.
\end{definition}

    To understand how arc consistency on a costgcc is established, some concepts from graph theory and flow theory are required.

\paragraph*{Graph theory:}

A \mev{directed graph} or \mev{digraph} $G=(X,U)$ consists of a
\mev{node set} $X$ and an \mev{arc set} $U$, where every arc $(x, y)$ 
is an ordered pair of distinct nodes. We will denote by $X(G)$ the
node set of $G$ and by $U(G)$ the arc set of $G$. The \mev{cost} of an arc is a value associated with the arc. When costs are associated with arcs, one should talk about weighted directed graphs.

A \mev{path} from node $x_1$ to node $x_k$ in $G$ is a list of nodes $[x_1, \dots, x_k]$ such that $(x_i, x_{i+1})$ is an arc for $i \in [1,k-1]$. 
The path is called \mev{simple} if all its nodes are distinct. 
The \mev{cost} of a path $P$, denoted by $cost(P)$, is the sum of the costs of the arcs contained in $P$. A \mev{shortest path} from a node $s$ to a node $t$ is a path from $s$ to $t$ whose cost is minimum.

\paragraph*{Flow theory:}

Let $G$ be a digraph where each arc $(x, y)$ 
is associated with three information: $l_{xy}$ the lower bound capacity, $u_{xy}$ the upper bound capacity and $c_{xy}$ the cost of the arc.

A \mev{flow} in $G$ is a function $f$ satisfying the following two
conditions: 
\point For any arc $(x, y)$, $f_{xy}$ represents the amount
of some commodity that can ``flow'' through the arc. Such a flow is
permitted only in the indicated direction of the arc, i.e., from $x$ to
$y$. For convenience, we assume $f_{xy}=0$ if $(x, y) \not\in
U(G)$.
\point A \mev{conservation law} is fulfilled at each
node: $\forall y \in X(G) : \sum_{x}f_{xy} = \sum_{z}f_{yz}$.

The \mev{cost} of a flow $f$ is $cost(f)=\sum_{(x, y) \in U(G)}f_{xy}c_{xy}$.

The feasible flow problem consists in computing a flow in $G$ that satisfies the \mev{capacity constraint}. That is finding $f$ such that $\forall (x, y) \in U(G)$ $l_{xy} \leq f_{xy} \leq u_{xy}$. The minimum cost flow problem consists in finding a feasible flow $f$ such that $cost(f)$ is minimum.

A min cost flow can be computed thanks to the augmenting shortest path algorithm. 
The main idea of the basic algorithms of flow theory is to proceed by
successive improvement of flows that are computed in a graph in which all
the lower bounds are zero and the
current flow is the zero flow (i.e., the flow value is zero on all arcs). 

First, assume that there is no lower capacity.
So, consider that all the lower bounds are equal to zero
and suppose that you want to increase the flow value for an arc $(x, y)$.
In this case, the zero flow is a feasible flow. 
Let $P$ be a shortest path from $y$ to
$x$ different from $(y, x)$, and $val=min(\{u_{xy}\} \cup \{u_{ab}$ s.t. $(a, b) \in P \})$. Then we can define the function $f$ on the arcs
of $G$ such that
$f_{ab}=val$ if $(a, b) \in P$ or $(a, b)=(x, y)$, and $f_{ab}=0$ otherwise. This
function is a flow in $G$ and $f_{xy} >0$. Now, from this flow we can define a particular graph without any flow value and all lower bounds equal to zero, the residual graph.

\begin{definition}
The \textbf{residual graph} for a given flow $f$, denoted by
$R(f)$, is the digraph with the same node set as in $G$ and with the arc set defined as follows:\\ $\forall (x, y) \in U(G)$:
\point $f_{xy} < u_{xy} \Leftrightarrow (x, y) \in U(R(f))$ and has cost
$rc_{xy}=c_{xy}$ and upper bound capacity
$r_{xy}=u_{xy} - f_{xy}$. 
\point $f_{xy} > l_{xy} \Leftrightarrow (y, x) \in U(R(f))$ and has cost
$rc_{yx}=-c_{xy}$ and upper bound capacity
$r_{yx}=f_{xy}-l_{xy}$. \\
All the lower bound capacities are equal to $0$.
\end{definition}

Then, we can select an arc and apply the previous algorithm on this arc in order to increase its flow value.
By dealing only with shortest path we can guarantee that the computed flow will have a minimum cost. 

Now consider the lower capacities. In this case, we can use the algorithm mentioned by Régin:

Start with the zero flow $f^o$. This flow satisfies the upper bounds.
Set $f=f^o$, and apply the following process while the flow is not feasible:\\
$1)$ pick an arc $(x, y)$ such that $f_{xy}$ violates the lower bound capacity in
$G$ (i.e., $f_{xy} < l_{xy}$).\\
$2)$ Find $P$ a shortest path from
$y$ to $x$ in $R(f)-\{(y, x)\}$.\\ 
$3)$ Obtain $f'$ from  $f$ by sending flow along $P$; set
$f=f'$ and goto $1)$\\
If, at some point, there is no
path for the current flow, then a feasible flow does not
exist. Otherwise, the obtained flow is feasible and is a minimum cost flow.

\section{Filtering Algorithm}
\label{sec:FilteringAlgorithm}
Our work builds on top of the original costgcc filtering (i.e., ~\cite{Regin:CostbasedArcConsistencyforGlobalCardinalityConstraints}). Before presenting how we speed up the algorithm for costgcc, let us briefly review the original algorithm.

There is a relation between a costgcc and the search for min-cost flow in a particular graph.

    \begin{definition}[\cite{Regin:CostbasedArcConsistencyforGlobalCardinalityConstraints}]
        Given $C = costgcc(X, l, u, cost, H)$. The value graph of $C$ is the bipartite graph $GV(C) = (X(C), D(X(C)), U)$ where $(x, a) \in U
        $ if $a \in D_x$. 
        The {\bf value network} of $C$ is the directed graph $N(C)$ with $l_{xy}$ the lower bound capacity, $u_{xy}$ the upper bound capacity and $c_{xy}$ the cost on arc from the node $x$ to the node $y$. $N(C)$ is obtained from the value graph $GV(C)$ by:
        \begin{itemize}
            \item Orienting each edge of $GV(C)$ from values to variables. $\forall x \in X(C) : \forall a \in D(x): l_{ax} =0$, $u_{ax} =1$ and $c_{ax} =cost(x,a)$.
            \item Adding a node $s$ and an arc from $s$ to each value. $\forall a \in D(X(C))$: $l_{sa} = l_a$, $u_{sa} = u_a$ and $c_{sa} =0$.
            \item Adding a node $t$ and an arc from each variable to $t$. $\forall x \in X(C): l_{xt} =1$, $u_{xt} =1$ and $c_{xt} =0$.
            \item Adding an arc $(t, s)$ with $l_{ts} \!\! = u_{ts} \!\! = \!\! |X(C)|$ and $c_{ts} \!\! = \!\! 0$.
        \end{itemize}
    \end{definition}

\begin{property}[\cite{Regin:CostbasedArcConsistencyforGlobalCardinalityConstraints}]
A costgcc $C$ is consistent iff there is a minimum cost flow in the value network of $C$ whose cost is less than or equal to H.
\end{property}

\begin{figure}
    \centering
    \begin{tikzpicture}[
    round/.style={regular polygon, regular polygon sides=6, draw=black},
    square/.style={rectangle, draw=black},
    oval/.style={rounded rectangle, draw=black, minimum width=1.5cm, minimum height=0.7cm},
    flecheG/.style={{Stealth[scale=1.3]}-},
    flecheD/.style={-{Stealth[scale=1.3]}},
    align=center
    every node/.style={transform shape}
    ]
    \node[round] (xt)      {t};
    
    \node[oval] (x1) [draw=red, right of=xt, yshift=3cm, xshift= 1.5cm, thick] {Peter};
    \node[oval] (x2) [below of=x1] {Paul};
    \node[oval] (x3) [below of=x2] {Mary};
    \node[oval] (x4) [below of=x3] {John};
    \node[oval] (x5) [below of=x4] {Bob};
    \node[oval] (x6) [below of=x5] {Mike};
    \node[oval] (x7) [below of=x6, draw=blue, thick] {Julia};

    \node (w1) [right of=x1, yshift=0.1cm, xshift= -0.1cm] {\textcolor{red}{-1}};
    \node (w2) [right of=x1, yshift=-0.8cm, xshift= -0.2cm] {\textcolor{red}{4}};
    \node (w3) [right of=x2, yshift=-0.2cm, xshift= 0.15cm] {-1};
    \node (w4) [right of=x2, yshift=-0.5cm, xshift= -0.2cm] {4};
    \node (w5) [right of=x3, yshift=0.17cm, xshift= 0.1cm] {\textcolor{red}{3}};
    \node (w6) [right of=x3, yshift=-0.2cm, xshift= -0.2cm] {\textcolor{red}{-1}};
    \node (w7) [right of=x4, yshift=0.54cm, xshift= 0.1cm] {3};
    \node (w8) [right of=x4, yshift=-0cm, xshift= -0.2cm] {-1};
    \node (w9) [right of=x5, yshift=0.5cm, xshift= -0.2cm] {-1};
    \node (w10) [right of=x6, yshift=0.5cm, xshift= -0.2cm] {-1};
    \node (w11) [right of=x7, yshift=0.8cm, xshift= -0.2cm] {\textcolor{blue}{-1}};
    \node (w12) [right of=x7, yshift=-0cm, xshift= -0.2cm] {\textcolor{blue}{1}};      

    \node[square] (x8) [right of=x1, yshift=-1cm, xshift= 2cm] {A};
    \node[square] (x9) [draw=red, below of=x8, thick] {B};
    \node[square] (x10) [below of=x9] {C};
    \node[square] (x11) [below of=x10] {D};
    \node[square] (x12) [below of=x11, draw=blue, thick] {E};

    \node (c1) [right of=x8, yshift=0.1cm, xshift= -0.2cm] {[1, 2]};
    \node (c2) [right of=x9, yshift=0.1cm, xshift= -0.2cm] {[1, 2]};
    \node (c3) [right of=x10, yshift=0.2cm, xshift= -0.2cm] {[1, 1]};
    \node (c4) [right of=x11, yshift=-0.1cm, xshift= -0.2cm] {[0, 2]};
    \node (c5) [right of=x12, yshift=-0.1cm,xshift= -0.2cm] {[0, 2]};

    \node[round] (xs) [right of=x8, yshift=-2cm, xshift= 1.5cm]      {s};
    

    \draw[flecheG] (x1.west) -- (xt);
    \draw[flecheG] (x2.west) -- (xt);
    \draw[flecheG] (x3.west) -- (xt);
    \draw[flecheG] (x4.west) -- (xt);
    \draw[flecheG] (x5.west) -- (xt);
    \draw[flecheG] (x6.west) -- (xt);
    \draw[flecheG] (x7.west) -- (xt);

    \draw[flecheG] (xs) -- (x8.east);
    \draw[flecheG] (xs) -- (x9.east);
    \draw[flecheG] (xs) -- (x10.east);
    \draw[flecheG, blue, thick] (xs) -- (x11.east);
    \draw[flecheD, blue, thick] (xs) -- (x12.east);

    \draw[flecheG, red, thick] (x8) -- (x1);
    \draw[flecheG] (x8) -- (x2);
    \draw[flecheD, red, thick] (x8) -- (x3);
    \draw[flecheD] (x8) -- (x4);

    \draw[flecheD, red, dashed, thick] (x9) -- (x1);
    \draw[flecheD] (x9) -- (x2);
    \draw[flecheG, red, thick] (x9) -- (x3);
    \draw[flecheG] (x9) -- (x4);

    \draw[flecheG] (x10) -- (x5);

    \draw[flecheG] (x11) -- (x6);
    \draw[flecheG, blue, thick] (x11) -- (x7);
    \draw[flecheD, blue, thick] (x12) -- (x7);

    \end{tikzpicture} 
    \caption{Example of computation of the consistency for the arcs ($E$, $Julia$) and ($B$, $Peter$). The value $B$ is not consistent with $Peter$. Thus, the dotted arc can be removed from the graph.}
    \label{fig:exempleArcConsistance}
\end{figure}
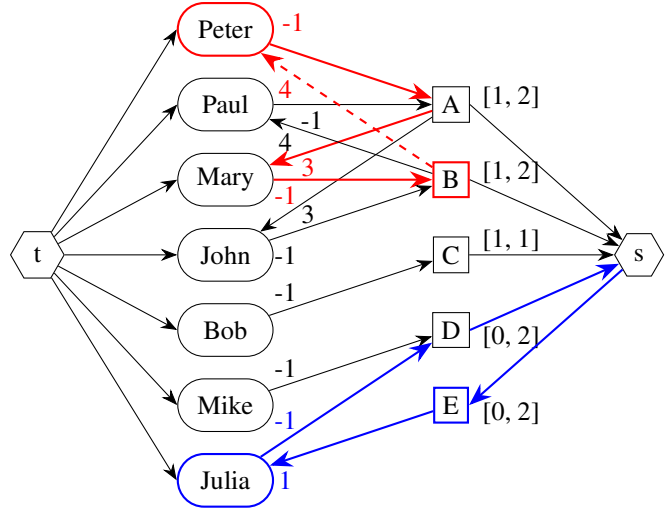

Figure~\ref{fig:exempleArcConsistance} represents the residual graph of the value network of the costgcc constraint defined in Figure~\ref{fig:intro}. 
This is the graph computed from a flow resulting of the min cost flow algorithm applied on the value network.
In the residual graph, the optimal solution corresponds to the arcs oriented from the variables to the values.
The optimal cost value is $7$.

For clarity, in the remainder, we consider that $C=costgcc(X, l, u, cost, H)$ is a costgcc constraint and that $f$ is min cost flow in $N(C)$. We also assume that the arc consistency of the underlined gcc of $C$ has been established.

The consistency of a value relates to the existence of a particular path in the residual graph of the min cost flow.
\begin{property}[\cite{Regin:CostbasedArcConsistencyforGlobalCardinalityConstraints}]\label{acxpte}
    A value $a$ of a variable $x$ is not consistent with $C$ iff the two following properties hold:
    \begin{itemize}
        \item $f_{ax} = 0$
        \item $d_{R(f)}(x, a) > H - cost(f) - rc_{ax}$
    \end{itemize}
    where $d_{R(f)}(x, a)$ is the shortest path between $x$ and $a$ in the residual graph of $f$, and $rc_{ax}$ is the residual cost of the arc $(a, x)$.
    \label{property:Regin}
\end{property}

To establish arc consistency, the previous property could be checked for each value of each variable. However it is possible to reduce the number of computed shortest paths.
    \begin{corollary}[\cite{Regin:CostbasedArcConsistencyforGlobalCardinalityConstraints}]\label{acpte}
        Given any variable $x$ and $b$ the value of $x$ such that $f_{bx} = 1$. Then, the value $a$ of $x$ is not consistent with $C$ iff the two following properties hold:
        \begin{itemize}
            \item $f_{ax} = 0$
            \item $d_{R(f)} (b, a) > H - cost(f) - rc_{ax} - rc_{xb}$
        \end{itemize}
    \end{corollary}

    An example of the application of Property~\ref{property:Regin} is given in Figure~\ref{fig:exempleArcConsistance}.
    The length of the shortest path from $Julia$ to $E$ has a cost of $-1$ (see blue arcs) and the cost of the arc ($E$, $Julia$) is $rc_{EJulia}=1$. Thus we have $d_{R(f)}(Julia, E)=-1$ and $H - cost(f) - rc_{EJulia}=11 - 7 -1=3$, so we have $-1 \leq 3$. From Property~\ref{property:Regin} it means that $(E, Julia)$ is consistent.
    The shortest path from $Peter$ to $B$ is $d_{R(f)}(Peter, B)=1$ and the cost of the arc ($B$, $Peter$) is $rc_{BPeter}=4$ (see red arcs). Hence, we have $H - cost(f) - rc_{BPeter} = 11 - 7 - 4 = 0$, so $1 > 0$. ($B$, $Peter$) is inconsistent, the arc is then removed.

\section{Upper Bounds of Shortest Paths}
\label{sec:UpperBounds}

Although Corollary~\ref{acpte} reduces the number of computations required to establish the arc consistency of the constraint, it systematically computes a large number of shortest paths. 
Precisely, the algorithm involves computing the shortest path between each assigned value and all other values which makes it difficult to use in practice. 
In addition, the constraint is often arc consistent, rendering any computation useless.
The aim of our approach is therefore to reduce the number of operations computed unnecessarily. 

We present a much more applied approach, based on the fact that Corollary~\ref{acpte} relies on the existence of a path of length less than a given value. It is not necessary to know the path precisely, or even to know its value. An upper bound is sufficient.

We can therefore immediately establish the following proposition:
    \begin{proposition}
         Let $B^+(x, a) \geq d_{R(f)}(x,a)$ be any upper bound on the length of the shortest path from $x$ to $a$. If 
            \[B^+(x, a) \leq H-cost(f)-rc_{ax}\]
        then the value $a$ of a variable $x$ is consistent with $C$.
    \end{proposition}

A good way of obtaining an upper bound on a distance between two points is to use the triangle inequality.
Here we are talking about the triangle inequality with respect to the shortest path distances in the graph, not an embedding in Euclidean space or some other metric, which need not be present.
    \begin{property}
        Let $x$, $y$, and $p$ be three nodes of a graph. According to the triangle inequality computed on shortest paths, we have:
        \[d(x, p) + d(p, y) \geq d(x, y)\]
        \label{property:triangleInequality}
        Here, $p$ is a particular node called landmark.
    \end{property}

Upper bounds obtained by the triangular inequality have been shown to be useful for guiding the computation of shortest paths. The ALT algorithm, yielding excellent results in practice for computing shortest paths in a very large graph, is based on this technique~\cite{Goldberg:ComputingtheShortestPath:ASearchMeetsGraphTheory}. 
    
Property~\ref{acxpte} and Corollary~\ref{acpte} can be rewritten for landmarks:
\begin{proposition}        \label{proposition:consistentArcWithHub}
    Given any variable $x$ such that $f_{bx} = 1$, $a$ any value of $x$ and $p$ any landmark. If one of the two condition is satisfied
   \[d_{R(f)}(x, p) + d_{R(f)}(p, a) \leq H - cost(f) - rc_{ax}\] 
    \[d_{R(f)}(b, p) + d_{R(f)} (p, a) \leq H - cost(f) - rc_{ax} - rc_{xb}\]
    
    then the value $a$ of $x$ is consistent with $C$.
\end{proposition}

The residual graph may have several strongly connected components. Each component must be treated separately. Thus, at least one landmark per component must be selected.

Thanks to the use of upper bounds we can go even further. It is possible to compute the consistency of all values of variables of a strongly connected component by checking a single condition.

\begin{definition}
    Consider $S$ a strongly connected component of $R(f)$, $p$ a landmark in $S$, $x \in S$ a variable, and $a$ a value of $x$. We define:
\begin{itemize}
    \item $d^{max}_{R(f)}(\cdot, p)=\max_{x \in S}(d_{R(f)}(x, p))$ 
    \item $d^{max}_{R(f)}(p, \cdot)=\max_{x \in S}(d_{R(f)}(p, x))$
    \item $rc^{max}=\max_{x \in S, a \in D(x)}(rc_{ax})$ 
\end{itemize}
\end{definition}

This leads to the following proposition:
\begin{proposition}        \label{proposition:formulemagique}
        Given $S$ a strongly connected component of $R(f)$ and $p$ a landmark in $S$. If 
        $$d^{max}_{R(f)}(\cdot, p) + d^{max}_{R(f)}(p, \cdot) \leq H - cost(f) - rc^{max}$$
        then all the values of all the variables involved in $S$ are consistent with $C$.
    \end{proposition}

The advantage of this method is that if the condition is satisfied, we can guarantee that all the values of a strongly connected component are consistent by computing only two shortest paths per landmark.

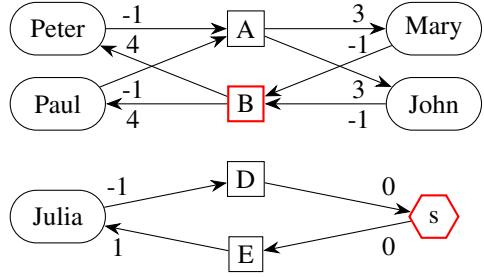
\begin{figure}
    \centering
    \begin{tikzpicture}[
        round/.style={regular polygon, regular polygon sides=6, draw=black},
        square/.style={rectangle, draw=black},
        oval/.style={rounded rectangle, draw=black, minimum width=1.5cm, minimum height=0.7cm},
        flecheG/.style={{Stealth[scale=1.3]}-},
        flecheD/.style={-{Stealth[scale=1.3]}},
        align=center
        every node/.style={transform shape}
        ]
    
    \node[oval] (x1) [] {Peter};
    \node[oval] (x2) [below of=x1] {Paul};
    \node[oval] (x7) [below of=x2, yshift=-0.5cm] {Julia};

    \node[square] (x8) [right of=x1, yshift=0cm, xshift= 1.5cm] {A};
    \node[square] (x9) [draw=red, below of=x8, thick] {B};
    \node[square] (x11) [below of=x9] {D};
    \node[square] (x12) [below of=x11] {E};

    \node[oval] (x3) [right of=x8, yshift=0cm, xshift= 1.5cm] {Mary};
    \node[oval] (x4) [below of=x3] {John};
    \node[round] (xs) [draw=red, below of=x4, yshift=-0.5cm, xshift= 0cm, thick]      {s};

    \node (w1) [right of=x1, yshift=0.2cm, xshift= -0cm] {-1};
    \node (w2) [right of=x1, yshift=-0.2cm, xshift= -0cm] {4};
    \node (w3) [right of=x2, yshift=0.2cm, xshift= -0cm] {-1};
    \node (w4) [right of=x2, yshift=-0.2cm, xshift= -0cm] {4};
    
    \node (w5) [left of=x3, yshift=0.2cm, xshift= 0cm] {3};
    \node (w6) [left of=x3, yshift=-0.2cm, xshift= -0cm] {-1};
    \node (w7) [left of=x4, yshift=0.2cm, xshift= 0cm] {3};
    \node (w8) [left of=x4, yshift=-0.2cm, xshift= -0cm] {-1};
    
    \node (w11) [right of=x7, yshift=0.4cm, xshift= -0.2cm] {-1};
    \node (w12) [right of=x7, yshift=-0.4cm, xshift= -0.2cm] {1};
    \node (w13) [left of=xs, yshift=0.4cm, xshift= 0.4cm] {0};
    \node (w14) [left of=xs, yshift=-0.4cm, xshift= 0.4cm] {0};


    \draw[flecheG] (xs) -- (x11);
    \draw[flecheD] (xs) -- (x12);

    \draw[flecheG] (x8) -- (x1);
    \draw[flecheG] (x8) -- (x2);
    \draw[flecheD] (x8) -- (x3);
    \draw[flecheD] (x8) -- (x4);

    \draw[flecheD] (x9) -- (x1);
    \draw[flecheD] (x9) -- (x2);
    \draw[flecheG] (x9) -- (x3);
    \draw[flecheG] (x9) -- (x4);

    \draw[flecheG] (x11) -- (x7);
    \draw[flecheD] (x12) -- (x7);

    \end{tikzpicture} 

    \caption{Example of landmark use. Nodes $B$ and $s$, shown in red, are selected as landmarks.}
    \label{fig:exampleLandmark}
\end{figure}

Figure~\ref{fig:exampleLandmark} gives an example of a residual graph on which Proposition~\ref{proposition:consistentArcWithHub} or~\ref{proposition:formulemagique} can be applied. There are 2 strongly connected components $\{Peter, A, Mary, Paul, B, John\}$ and $\{Julia, D, s, E\}$. At least 2 landmarks are required (one for each component). We select $B$ and $s$ arbitrarily. 

Thanks to Proposition~\ref{proposition:formulemagique} we see that the maximum shortest path through $s$ is the path from $D$ to $Julia$ with $d^{max}_{R(f)}(\cdot, S)=d(D, s)=0$ and $d^{max}_{R(f)}(s, \cdot)=d(s, Julia)=1$. Furthermore, the longest arc of this strongly connected component is $rc^{max}=rc_{EJulia}=1$. Thus we have $d(D, s)+d(s, Julia)=1$ and $H-cost(f) - rc_{EJulia}=3$, so we have $1 \leq 3$. This confirms that all the values of variables in the strongly connected component of $s$ are consistent with the constraint.
If the Proposition~\ref{proposition:formulemagique} can guarantee that all values of variables are consistent in this strongly connected component then we can easily deduce that the Proposition~\ref{proposition:consistentArcWithHub} can also do it.

For the first strongly connected component, Proposition~\ref{proposition:consistentArcWithHub} and~\ref{proposition:formulemagique} do not guarantee the consistency 
of the values of the variables. It is therefore necessary to compute exact shortest paths between values and variables.

\section{Improved Filtering Algorithm}
\label{sec:ImprovedFilteringAlgorithm}
We can now describe Algorithm~\ref{alg:algorithm}, which eliminates values that are inconsistent with the constraint. The algorithm takes as parameters a min cost flow $f$, its residual graph $R(f)$, a strongly connected component represented by its set of nodes $S$ and $P$ a set of landmarks of $S$. This algorithm must therefore be called for each strongly connected component. The algorithm begins by checking whether Proposition~\ref{proposition:formulemagique} holds. If true, then the algorithm stops, since this means that all the values of the variables in the connected component $S$ are consistent.
Otherwise, it is necessary to check each value potentially inconsistent individually.
So, for each of those values Proposition~\ref{proposition:consistentArcWithHub} is checked. If it is satisfied, then the value is consistent, otherwise an explicit shortest path is computed to determine whether the value is consistent or not.

\begin{algorithm}[tb]
{\footnotesize
\fntitrealgo{arcConsistencyWithLandmarks}$(f,R_f,S,P)$\;
    \For{$p \in P$}{
        $d(p, \cdot) \leftarrow shortestPath_{R(f)}(p, \cdot)$ // shortest path in $R(f)$ \; 
        $d(\cdot, p) \leftarrow shortestPath_{\overline{R}(f)}(p, \cdot)$ // shortest path in $\overline{R}(f)$, the reverse graph of $R(f)$ \;
    }
    // Check of Proposition~\ref{proposition:formulemagique} \;
    $rc^{max} \leftarrow \max_{x \in S, a \in D(x)}(rc_{ax})$ \;
    \For{$p \in P$}{
    $d^{max}_{R(f)}(\cdot, p) \leftarrow \max_{x \in S}(d_{R(f)}(x, p))$ \; 
    $d^{max}_{R(f)}(p, \cdot) \leftarrow \max_{x \in S}(d_{R(f)}(p, x))$ \;
    \If{$d^{max}_{R(f)}(\cdot, p) + d^{max}_{R(f)}(p, \cdot) \leq H - cost(f) - rc^{max}$}{
        // all values of all variables of $S$ are consistent \;
        return \;
    }
    }
    $\Delta \leftarrow \{a$ such that $f_{sa} > 0\}$ \;
    \For{value b $\in \Delta$}{
        \For{$x$ such that $f_{bx} =1$}{
            $\delta (b) \leftarrow \{a$ such that $a \in D(x)$ and $a \neq b\}$ \;
            $computePath \leftarrow false$ \;
            // Check of Proposition~\ref{proposition:consistentArcWithHub} \;
            \For{$a \in D(x)$ {\bf while not} $computePath$}{
                $dpmin \leftarrow \min_{p \in P}(d_{R(f)}(x, p) + d_{R(f)}(p, a))$\;
                \If{ $dpmin > H - cost(f) - rc_{ax}$}{$computePath \leftarrow true$ \;}
            }
            // Check for an explicit shortest path computation \;
            \If{$computePath$}{
                $d_{R(f)}(b, \cdot) \leftarrow shortestPath(b, \cdot)$ \;
                \For{$a \in D(x)$}{
                    \lIf{$d_{R(f)}(b, a) > H - cost(f) - rc_{ax} - rc_{xb}$}{
                        remove $a$ from $D(x)$}
                }
            }
        }
    }
    }
    \caption{Arc Consistency Algorithm for a Strongly Connected Component}
    \label{alg:algorithm}
\end{algorithm}

When testing Corollary \ref{acpte}, we could refine the algorithm by identifying the nodes for which we need to search for a shortest path from $b$ to them, but this is not interesting in practice as the shortest path algorithm will quickly find that they are at an acceptable distance from $b$. 

\paragraph*{Practical improvements:}

One can compute landmarks only when they are needed. This consideration is effective in practice and a simple modified version of the basic algorithm is possible. This modification proceeds by iteration over the landmarks. Consider $V$ the set of values for which a shortest path must be computed. 

The following process is defined:
The landmark $p$ is considered. 
Proposition~\ref{proposition:formulemagique} is checked according to $p$. If it is satisfied then $V$ is emptied (all values are consistent) otherwise the values $V$ that satisfies Proposition~\ref{proposition:consistentArcWithHub} according to $p$ are removed from $V$, because they are consistent.

This process is repeated while $V$ is not empty and some landmarks remain. 
In other words, the landmarks are successively considered while the status of some values is not determined.

If there are no more landmarks to compute, then, and only then, shortest paths are explicitly computed for the value in $V$. 
In practice, it is frequent to find that all values are consistent without using all the landmarks.
This practical improvement means that not all landmarks need to be systematically computed.

Note that the landmark approach subsumes all the practical improvements proposed by Régin.

As far as the shortest path algorithm is concerned, it is interesting to remove the negative costs from the residual graph in order to use Dijkstra's algorithm, as mentioned by Régin. It only requires one shortest path computation~\cite{Regin:CostbasedArcConsistencyforGlobalCardinalityConstraints}. 

\paragraph*{Complexity:}

Let $SP$ be the complexity of computing a shortest path from one node to all others.
Régin's algorithm has a complexity of $\Omega(\delta \times SP)$ in the best case and $O(\delta \times SP)$ in the worst case, where $\delta$ is the number of assigned values.
With landmarks, the complexity in the best case is in $\Omega(FindP + |P| \times SP)$ where $|P|$ is the number of landmarks and $FindP$ is the complexity of finding the landmarks. This complexity is obtained when Proposition~\ref{proposition:formulemagique} detects that every value is consistent. Note that, this detection can happen on the first landmark and so we can have $|P|=1$. In the worst case, the complexity is the same as that of Régin's algorithm, provided that $|P|$ is in $O(\delta)$ and $FindP$ is in $O(\delta \times SP)$.
As with the ALT method, we consider several landmarks in order to have a better chance of finding landmarks that avoid explicit shortest path computations.

\section{Landmark Selection}
\label{sec:LandmarkSelection}

There are different methods for selecting landmarks.
    \paragraph*{Random:}         
        A landmark is randomly selected. This method is fast to find landmarks, so we used it to compare to other methods. 
        
    \paragraph*{Outline:}
    The method is based on an approximation of the outlines of a graph.
        \begin{definition}
            The \textbf{outlines of a graph} G are defined by one or more pairs of nodes $(x, y)$ with $x, y \in X$ that maximize the minimum distance between $x$ and $y$ among all pairs of nodes in the graph.
        \end{definition}
To find the pair of nodes representing the outline, we use a well-known 2-approximation. First, we perform a shortest-path search starting from an arbitrary node $x$, then select the node $y$, which is the furthest node from $x$, as a landmark. Next, the shortest paths from $y$ are computed, and $z$ the node furthest from $y$ is selected as the second landmark. The outline is therefore ($y$, $z$) and the landmarks $y$ and $z$.
The complexity of finding a landmark depends on the complexity of computing the two shortest paths, and is therefore in $O(SP)$.        
    \paragraph*{Center:}
        The method is based on an approximation of the center of a graph.
        \begin{definition}
            The \textbf{center of a graph} G is defined by one or more nodes $x \in X$ that minimize the maximum distance from them to any other node in the graph.
        \end{definition}
As the definition of the outlines and the center are 
similar, the selection of landmarks is also similar. We search for the outlines $(x, y)$ with $x, y \in X$ and select as the center the node $z$ that lies halfway between $x$ and $y$. The landmark is $z$. 
The complexity is the same as for the previous method, $O(SP)$.

    \paragraph*{Outline \& center:}
The method is based on both outlines and center of a graph, that is a pair of outlines and a center are selected as described earlier. 

    \paragraph*{Maximum degree:}
        The method is based on the node's degree. We select as a landmark the node $x \in X$ that maximizes $(deg^+(x) + deg^-(x)) \times min(deg^+(x), deg^-(x))$, where $deg^+(x)$ (resp. $deg^-(x)$) is the number of outgoing arcs of $x$ (resp. incoming arcs to $x$). We used this formula to choose nodes with a large number of predecessors and successors. We also expect to choose nodes with a good balance between predecessors and successors. 
        To find landmarks we traverse every node once, giving a complexity of $O(|X|)$. 
        
    All these methods must be applied for each strongly connected component.

\section{Experimentation} 
\label{sec:Experimentation}

The experiments were carried out on a computer with an Intel Core i7-3930K CPU 3,20 GHz processor, 64 GB of memory and running under Windows 10 Pro.
All algorithms were implemented in Java (openjdk-17) in an internal CP solver.

The results relate to the solving of four problems, 
the traveling salesman problem (TSP)~\cite{Isoart:Thetravelingsalesmanprobleminconstraintprogramming}, the StockingCost problem~\cite{Houndji:dataset_item}, the flexible job shop scheduling problem (FJSSP)~\cite{Pelleau:dataFJSSP,Weise:jsspInstancesAndResults} and a problem of assigning child to activities (CHILD)~\cite{Varone:dataCHILD}. The TSP data are the instances (77) of the TSPLIB~\cite{Reinelt:TSPLIB} having less than 1,500 cities. Some of them involve more than a million of edges. 
The StockingCost data are those used in a Houndji's paper~\cite{Houndji:dataset_item}, this is random data distributed define as 
100 instances with 500 periods. Precisely, 
the StockingCost instances have 500 variables and 500 values. 
The FJSSP data come from two different sources, given 
by Pelleau~\cite{Pelleau:dataFJSSP} and Weise~\cite{Weise:jsspInstancesAndResults}. There are 370 instances with between 5 and 20 variables linked to a few values (between 5 and 10), and most instances have between 50 and 300 arcs. 
The CHILD instance contains only real-life data from~\cite{Varone:dataCHILD}. There are 623 children and 317 activities. Each child must be assigned to one activity. One activity can be associated with multiple children.

For each instance of each problem, we measure the information relating to the establishment of the arc consistency of the costgcc constraint at the root of the search tree. The mean of the results for each data set are reported in the tables.

The $H$ value of the TSP instances comes from the heuristic of Lin-Kernighan~\cite{Lin:HeuristicTSP}. Most of the time, this value is the optimal value. For the instances 
StockingCost, FJSSP and CHILD the regular $H$ is the smallest value such that there exists at least one solution and the big $H$ is twice as large as the regular $H$.

It is important to pay close attention to the relationship between the value of $H$ and the value of the minimum-cost flow. Indeed, the costgcc constraint sometimes represents a lower bound of the optimal solution, and this lower bound can be more or less distant from the optimal solution. 
So if $H$ is the value of the optimal solution, then the min-cost flow may well have a much lower value.  This is particularly true for the TSP problem.

Shortest paths are computed by using Dijkstra's algorithm and strongly connected components are computed by using Tarjan's algorithm.

The following abbreviations are used for the landmark selection methods: C for the center selection, O for the outline selection, C \& O for the combination between center and outline, Deg for the selection based on the maximum degree and R for the random selection. In addition, line 5+ contains the minimum values for a number of landmarks ranging from 5 to 10.

\pagebreak

\subsection{Results Tables}

\begin{table}[h!]
    \centering
    \resizebox{0.6\columnwidth}{!}{
    \begin{tabular}{| c | c || c | c | c | c | c | c |}
         \hline
         & Régin & \multirow{2}{1.3cm}{Landmark Number} & C & O & C \& O & Deg & R \\
          & & &   &   &   &   &   \\
         \hline
         \hline
         \multirow{5}{2.1cm}{TSP ($\le$ 100 cities)} 
            & \multirow{4}{0.8cm}{57.6}
                & 1 & 31.7 & 36.3 & 36.2 & \textbf{27.7} & \textbf{27.7} \\
                & & 2 & 35.3 & 39.9 & 39.8 & 32.5 & 29.5 \\
                & & 3 & 38 & 42.7 & 42.5 & 32.5 & 28.5 \\
                & & 4 & 41.6 & 46.3 & 46.1 & 32 & 30.1 \\
                & & 5+ & 44.8 & 50 & 50.2 & 32 & 32.2 \\
         \hline
         \multirow{5}{2.1cm}{TSP ($>$ 100 \& $<$ 250 cities)}
            & \multirow{4}{0.8cm}{163.3}
               & 1 & 42.2 & 45 & 47.9 & \textbf{40.5} & \textbf{40.5} \\
               & & 2 & 44.4 & 47.4 & 46.3 & 41.6 & 41.6 \\
               & & 3 & 46 & 49.3 & 48.2 & 41.2 & 41.2 \\
               & & 4 & 48.6 & 51.9 & 50.8 & 42.3 & 42.3 \\
               & & 5+ & 50.2 & 54.1 & 52.2 & 43.1 & 43.3 \\
         \hline
         \multirow{5}{2.1cm}{TSP ($\ge$ 250 cities)}
           & \multirow{4}{0.8cm}{662.7}
               & 1 & 18.1 & 19.8 & 19.8 & 17.8 & 17.8 \\
               & & 2 & 18.5 & 21.4 & 21.4 & 18.1 & 18.1 \\
               & & 3 & 18.5 & 21 & 19.3 & \textbf{16.3} & \textbf{16.2} \\
               & & 4 & 18.8 & 21.6 & 19.9 & 16.4 & \textbf{16.3} \\
               & & 5+ & 19 & 21.8 & 20.1 & 16.7 & 16.4 \\
         \hline
         \multirow{5}{2.1cm}{StockingCost (Regular H)} 
           & \multirow{4}{0.8cm}{\textbf{493.3}}
               & 1 & 496.9 & 497.3 & 496.9 & 495.3 & 495.3 \\
               & & 2 & 500.8 & 501.2 & 500.8 & 497.3 & 497.2 \\
               & & 3 & 504.7 & 505.1 & 504.7 & 499.2 & 499.1 \\
               & & 4 & 508.6 & 509 & 508.6 & 501.2 & 501 \\
               & & 5+ & 512.5 & 512.9 & 512.6 & 503.2 & 503 \\
         \hline
         \multirow{5}{2.1cm}{StockingCost (Big H)} 
           & \multirow{4}{0.8cm}{493.3}
               & 1 & 4 & 4 & 4 & \textbf{2} & \textbf{2} \\
               & & 2 & 4 & 4 & 4 & \textbf{2} & \textbf{2} \\
               & & 3 & 4 & 4 & 4 & \textbf{2} & \textbf{2} \\
               & & 4 & 4 & 4 & 4 & \textbf{2} & \textbf{2} \\
               & & 5+ & 4 & 4 & 4 & \textbf{2} & \textbf{2} \\
         \hline
         \multirow{5}{2.1cm}{FJSSP (Regular H)} 
           & \multirow{4}{0.8cm}{10.4}
               & 1 & 8.3 & 5.1 & 4.8 & \textbf{2} & 6.3 \\
               & & 2 & 8.3 & 5.1 & 4.8 & \textbf{2} & 5.3 \\
               & & 3 & 8.3 & 5.1 & 4.8 & \textbf{2} & 4.6 \\
               & & 4 & 8.3 & 5.1 & 4.8 & \textbf{2} & 4 \\
               & & 5+ & 8.3 & 5.1 & 4.8 & \textbf{2} & 4 \\
         \hline
         \multirow{5}{2.1cm}{FJSSP (Big H)} 
           & \multirow{4}{0.8cm}{10.4}
               & 1 & 4.5 & 4.3 & 4.3 & \textbf{2} & 3.2 \\
               & & 2 & 4.5 & 4.3 & 4.3 & \textbf{2} & 2.8 \\
               & & 3 & 4.5 & 4.3 & 4.3 & \textbf{2} & 2.6 \\
               & & 4 & 2.9 & 4.3 & 4.3 & \textbf{2} & 2.4 \\
               & & 5+ & 2.9 & 4.3 & 4.3 & \textbf{2} & 2.4 \\
         \hline
         \multirow{5}{2.1cm}{CHILD (Regular H)} 
           & \multirow{4}{0.8cm}{\textbf{108}}
               & 1 & 112 & 112 & 112 & 109 & 110 \\
               & & 2 & 116 & 116 & 116 & 111 & 112 \\
               & & 3 & 120 & 120 & 120 & 113 & 114 \\
               & & 4 & 124 & 124 & 124 & 115 & 116 \\
               & & 5+ & 128 & 128 & 128 & 117 & 118 \\
         \hline
         \multirow{5}{2.1cm}{CHILD (Big H)} 
           & \multirow{4}{0.8cm}{108}
               & 1 & 4 & 4 & 4 & \textbf{2} & \textbf{2} \\
               & & 2 & 4 & 4 & 4 & \textbf{2} & \textbf{2} \\
               & & 3 & 4 & 4 & 4 & \textbf{2} & \textbf{2} \\
               & & 4 & 4 & 4 & 4 & \textbf{2} & \textbf{2} \\
               & & 5+ & 4 & 4 & 4 & \textbf{2} & \textbf{2} \\
         \hline
    \end{tabular}
    }
    \caption{Establishment of the arc consistency of a costgcc constraint: average number of computed shortest paths
    depending on the number of landmarks and the landmark selection method.}
    \label{tab:numberOfShortestPath}
\end{table}

\begin{table}[!ht]
    \centering
    \resizebox{0.7\columnwidth}{!}{
    \begin{tabular}{| c || c | c | c | c | c | c |}
         \hline
         & Régin & C & O & C \& O & Deg & R \\
         \hline
         \hline
         TSP ($\le$ 100 cities)
              & 35.8 & 19.7 & 24.4 & 24.3 & \textbf{8.3} & \textbf{8.2}  \\
               
         \hline
         TSP ($>$ 100 \& $<$ 250 cities) 
               & 131.1 & 16 & 19.7 & 18.6 & \textbf{10.1} & \textbf{10.1}  \\
               
         \hline
         TSP ($\ge$ 250 cities)
               & 649.8 & 5.9 & 8.6 & 6.9 & \textbf{3.4} & \textbf{3.3}  \\
               
         \hline
         StockingCost (Regular H)
                & \textbf{0} & 3.2 & 7.6 & 11.2 & 7.8 & 7.6 \\               
         \hline
         StockingCost (Big H)
                & 493.3 & 4 & 4 & 4 & \textbf{2} & \textbf{2} \\
               
         \hline
         FJSSP (Regular H)
                & \textbf{0} & 8.3 & 5.1 & 4.8 & 4 & 4 \\
               
         \hline
         FJSSP (Big H)
                & 10.1 & 2.9 & 4.3 & 4.3 & \textbf{2} & 2.4 \\
                
         \hline
         CHILD (Regular H)
               & \textbf{0} & 16 & 16 & 16 & 8 & 8 \\
               
         \hline
         CHILD (Big H)
                & 108 & 4 & 4 & 4 & \textbf{2} & \textbf{2} \\
                
         \hline
    \end{tabular}
    }
    \caption{Establishment of the arc consistency of a costgcc constraint: number of average shortest paths computed uselessly with 4 landmarks.}
    \label{tab:numberSPUseless}
\end{table}

\begin{table}[!ht]
    \centering
    \resizebox{0.7\columnwidth}{!}{
    \begin{tabular}{| c | c || c | c | c | c | c | c |}
         \hline
         & & Régin & C & O & C \& O & Deg & R \\
         \hline
         \hline
         \multirow{3}{2.1cm}{TSP ($\le$ 100 cities)} 
               & Mean & 7.3 & 5.9 & 6 & 6.6 & 5.7 & \textbf{4.5} \\
               & Median & 3.4 & 3.6 & 4.4 & 4.1 & 3.6 & \textbf{3.3} \\
               & Ratio & & 1.2 & 1.2 & 1.1 & 1.3 & \textbf{1.6} \\
         \hline
         \multirow{3}{2.1cm}{TSP ($>$ 100 \& $<$ 250 cities)} 
               & Mean & 76.6 & \textbf{29.8} & 30.6 & 30.2 & \textbf{28.6} & 31.1 \\
               & Median & 51.2 & \textbf{14.3} & 16 & 17 & 15.4 & \textbf{14.3} \\
               & Ratio & & \textbf{2.6} & 2.5 & 2.5 & \textbf{2.7} & 2.5 \\
         \hline
         \multirow{3}{2.1cm}{TSP ($\ge$ 250 cities)}
               & Mean & 12124.9 & 278.9 & 275.2 & 275.4 & \textbf{213} & 265 \\
               & Median & 2310.2 & 126.8 & 117.7 & 90.6 & 89.1 & \textbf{85.9} \\
               & Ratio & & 43.5 & 44.1 & 44 & \textbf{56.9} & 45.8 \\
         \hline
         \multirow{3}{2.1cm}{StockingCost (Regular H)} 
               & Mean & 603.83 & \textbf{511.8} & 617.9 & 626.2 & 580.3 & 639.4 \\
               & Median & 585.7 & 553.3 & 186.9 & 186.4 & 248 & \textbf{166.4} \\
               & Ratio & & \textbf{1.2} & 1 & 1 & 1 & 0.9 \\
         \hline
         \multirow{3}{2.1cm}{StockingCost (Big H)} 
               & Mean & 534.76 & 34.1 & 32.4 & \textbf{31.6} & 33.2 & 32.6 \\
               & Median & 519.1 & 33.8 & 32.4 & 31.9 & 32.8 & \textbf{30.1} \\
               & Ratio & & 15.7 & 16.5 & \textbf{16.9} & 16 & 16.4 \\
         \hline
         \multirow{3}{2.1cm}{FJSSP (Regular H)} 
               & Mean & 0.4 & 0.5 & \textbf{0.3} & 0.4 & 0.4 & 0.5 \\
               & Median & \textbf{0.1} & 0.3 & 0.2 & 0.3 & 0.2 & 0.3 \\
               & Ratio & & 0.8 & \textbf{1.7} & 0.75 & 1 & 0.8 \\
         \hline
         \multirow{3}{2.1cm}{FJSSP (Big H)} 
               & Mean & 0.4 & 0.4 & \textbf{0.3} & \textbf{0.3} & \textbf{0.3} & \textbf{0.3} \\
               & Median & \textbf{0.1} & 0.2 & 0.2 & 0.2 & 0.2 & 0.2 \\
               & Ratio & & 1 & \textbf{1.3} & \textbf{1.3} & \textbf{1.3} & \textbf{1.3} \\
         \hline
         \multirow{2}{2.1cm}{CHILD (Regular H)} 
               & Time & 65.1 & 69.2 & \textbf{54.4} & 67.6 & 75.9 & 65.4 \\
               & Ratio & & 0.9 & \textbf{1.2} & 1 & 0.8 & 1 \\
         \hline
         \multirow{2}{2.1cm}{CHILD (Big H)} 
               & Time & 58.2 & 7 & 6.5 & 7.3 & \textbf{6} & \textbf{6} \\
               & Ratio & & 8.3 & 9 & 8 & \textbf{9.7} & \textbf{9.7} \\
         \hline
    \end{tabular}
    }
    \caption{Establishment of the arc consistency of a costgcc constraint: computation times (in ms) and ratio. Experimentation with 4 landmarks.}
    \label{tab:timeWithMagic}
\end{table}

We consider a shortest path calculation to be the calculation of the shortest paths from one node to all the others. 

The number of shortest paths calculated is an important parameter for distinguishing between algorithms. 
Some shortest path computations cannot be avoided, particularly those required to detect inconsistent values. 
However, some shortest path computations are useless, as they do not allow us to establish the inconsistency of any value. Precisely, if the shortest path computation from $b$ In Corollary~\ref{acpte} does not lead to any deletion of values then this path computation is useless. 

Table~\ref{tab:numberOfShortestPath} compares the number of shortest path computations performed by Régin's algorithm and by our approach as a function of the number of landmarks allowed in. The number of shortest paths required to compute landmarks are included.

Table~\ref{tab:numberSPUseless} shows the average number of useless shortest path computations for each dataset. We consider that shortest path computations for landmarks are always useless, so they are always included. That is why there are never $0$ computations with landmarks. 

Table~\ref{tab:timeWithMagic} gives the time required by each method.

\subsection{Results Analysis}

Table~\ref{tab:numberOfShortestPath} shows that our approach generally computes significantly fewer shortest paths than Régin's algorithm for all landmarks selection methods. For the TSP instances, we compute on average between 2 and 47 times fewer shortest paths than Régin's algorithm. The difference is significant for all instances except for the StockingCost instances with Regular H.
It should be noted that our approach is always better or equivalent and allows us to detect quickly whether the constraint is arc consistent in certain cases.

In the best case, our approach does not compute any shortest paths other than those required to determine landmarks. Our approach can compute more shortest paths only when there is no inconsistent arc and the extra computation is due to the landmarks.
The number of useless path computations is also reduced by our method (See Table~\ref{tab:numberSPUseless}).

For computation times, we find the same kind of results as before (See Table~\ref{tab:timeWithMagic}). The gain average factors evolve between $1$ and $57$. 

\subsubsection{Landmark Number and Selection Method}

We can see that the results do not change much as a function of the number of landmarks.
The major part of problems have best or equivalent results with 4 landmarks, but the difference is minimal. When it is not mentioned $4$ landmarks are used. 

Two methods of landmark selection appear to be more effective in practice: the method based on maximum node degrees and the random node selection method. As there is little difference between these two methods, and the former is more robust than the latter, we recommend defining landmarks based on maximum degree nodes. 

\subsubsection{Impact of the practical improvement of Section 5}

Thanks to this practical improvement all the authorized landmarks are not systematically used. This is clearly seen for StockingCost and CHILD instances with big $H$. The computation of a single landmark is sufficient to guarantee that all the values are consistent.

\subsubsection{StockingCost, FJSSP and CHILD problems}

For the StockingCost, FJSSP and CHILD problems, the results strongly depends on the value of $H$.

For the Regular $H$ value the results are similar to those of Régin's algorithm. In these problems, Regular $H$ is close to the optimal value of the min cost flow of the underlined costgcc.
Thus, there is less margin and therefore more inconsistent values.
FJSSP instances are also small and do not allow us to highlight the usefulness of landmarks. Indeed, in a small instance, computing a landmark gives us access to less information than in a large instance. In addition, for practical use, it is more interesting to save time on large instances since they take longer to resolve than on small instances which are already quick to resolve.

For a Big $H$ value the landmark method clearly outperforms Régin's algorithm.

\begin{figure}[ht!]
    \begin{subfigure}[t]{0.33\columnwidth}
    \resizebox{1.1\columnwidth}{!}{
        \centering
        \begin{tikzpicture}
    \begin{axis}[
        xlabel={H multiplier},
        ylabel={Average number of removed arcs},
        xmin=1, xmax=2,
        ymin=0, ymax=11000,
        xtick={1, 1.1, 1.2, 1.3, 1.4, 1.5, 1.6, 1.7, 1.8, 1.9, 2},
        ytick={0, 2000, 4000, 6000, 8000, 10000, 11000},
        legend pos=north west,
        ymajorgrids=true,
        grid style=dashed,
    ]

    \addplot[
        color=blue,
        ]
        coordinates {
    (1, 10195.0) (1.1, 0.0) (1.2, 0.0) (1.3, 0.0) (1.4, 0.0) (1.5, 0.0) (1.6, 0.0) (1.7, 0.0) (1.8, 0.0) (1.9, 0.0) (2, 0.0)     };
        
    \end{axis}
    \end{tikzpicture}
    }
        \subcaption{CHILD}
    \end{subfigure}
    \quad
    \begin{subfigure}[t]{0.3\columnwidth}
    \resizebox{1.1\columnwidth}{!}{
        \centering
        \begin{tikzpicture}
    \begin{axis}[
        xlabel={H multiplier},
        xmin=1, xmax=2,
        ymin=0, ymax=110,
        xtick={1, 1.1, 1.2, 1.3, 1.4, 1.5, 1.6, 1.7, 1.8, 1.9, 2},
        ytick={0, 20, 40, 60, 80, 100, 110},
        legend pos=north west,
        ymajorgrids=true,
        grid style=dashed,
    ]

    \addplot[
        color=blue,
        ]
        coordinates {
    (1, 109.5) (1.1, 46.7) (1.2, 42.1) (1.3, 23.6) (1.4, 7.7) (1.5, 6.3) (1.6, 4.1) (1.7, 5) (1.8, 3.7) (1.9, 3.3) (2, 3.3)      };
        
    \end{axis}
    \end{tikzpicture}
    }
        \subcaption{FJSSP}
    \end{subfigure}
    \quad
    \begin{subfigure}[t]{0.3\columnwidth}
    \resizebox{1.1\columnwidth}{!}{
        \centering
        \begin{tikzpicture}
    \begin{axis}[
        xlabel={H multiplier},
        xmin=1, xmax=2,
        ymin=0, ymax=36000,
        xtick={1, 1.1, 1.2, 1.3, 1.4, 1.5, 1.6, 1.7, 1.8, 1.9, 2},
        ytick={0, 5000, 10000, 15000, 20000, 25000, 30000, 35000},
        legend pos=north west,
        ymajorgrids=true,
        grid style=dashed,
    ]

    \addplot[
        color=blue,
        ]
        coordinates {
    (1, 35410.1) (1.1, 2489.1) (1.2, 0.0) (1.3, 0.0) (1.4, 0.0) (1.5, 0.0) (1.6, 0.0) (1.7, 0.0) (1.8, 0.0) (1.9, 0.0) (2, 0.0)       };        
    \end{axis}
    \end{tikzpicture}
    }
        \subcaption{StockingCost}
        \end{subfigure}
    \caption{Evolution of the average number of removed arcs for the CHILD, FJSSP and StockingCost instances in function of the multiplier of $H$.}
    \label{fig:averageNumberOfRemovedArcInFunctionOfH}
\end{figure}
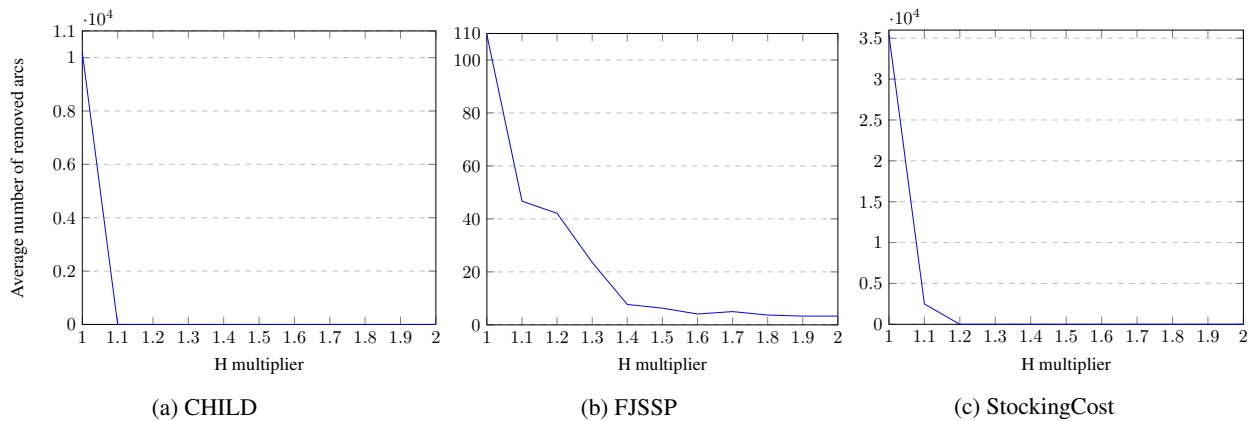

\begin{figure}[ht!]
    \begin{subfigure}[t]{0.33\columnwidth}
    \resizebox{1.1\columnwidth}{!}{
         \centering
        \begin{tikzpicture}
    \begin{axis}[
        xlabel={H multiplier},
        ylabel={Average number of useless path computations},
        xmin=1, xmax=2,
        ymin=0, ymax=110,
        xtick={1, 1.1, 1.2, 1.3, 1.4, 1.5, 1.6, 1.7, 1.8, 1.9, 2},
        ytick={0, 20, 40, 60, 80, 100, 110},
        legend style={xshift=2cm},
        legend pos=north west,
        ymajorgrids=true,
        grid style=dashed,
    ]

    \addplot[
        dashed,
        color=blue,
        ]
        coordinates {
    (1, 0.0) (1.1, 108.0) (1.2, 108.0) (1.3, 108.0) (1.4, 108.0) (1.5, 108.0) (1.6, 108.0) (1.7, 108.0) (1.8, 108.0) (1.9, 108.0) (2, 108.0)        };

    \addplot[
        color=red,
        ]
        coordinates {
    (1, 16.0) (1.1, 8.0) (1.2, 8.0) (1.3, 8.0) (1.4, 8.0) (1.5, 8.0) (1.6, 8.0) (1.7, 8.0) (1.8, 8.0) (1.9, 8.0) (2, 8.0) };

    \addplot[
        color=green,
        ]
        coordinates { 
    (1, 16.0) (1.1, 8.0) (1.2, 8.0) (1.3, 8.0) (1.4, 8.0) (1.5, 8.0) (1.6, 8.0) (1.7, 8.0) (1.8, 8.0) (1.9, 8.0) (2, 8.0)     };

    \addplot[
        ]
        coordinates {
    (1, 16.0) (1.1, 12.0) (1.2, 8.0) (1.3, 8.0) (1.4, 8.0) (1.5, 8.0) (1.6, 8.0) (1.7, 8.0) (1.8, 8.0) (1.9, 8.0) (2, 8.0) 
    };

    \addplot[
        color=orange,
        ]
        coordinates {
    (1, 8.0) (1.1, 2.0) (1.2, 2.0) (1.3, 2.0) (1.4, 2.0) (1.5, 2.0) (1.6, 2.0) (1.7, 2.0) (1.8, 2.0) (1.9, 2.0) (2, 2.0) };
    
    \addplot[
        color=purple,
        ]
        coordinates {
    (1, 8.0) (1.1, 2.0) (1.2, 2.0) (1.3, 2.0) (1.4, 2.0) (1.5, 2.0) (1.6, 2.0) (1.7, 2.0) (1.8, 2.0) (1.9, 2.0) (2, 2.0) 
    };
            
    \end{axis}
    \end{tikzpicture}
    }
        \subcaption{CHILD}
    \end{subfigure}
    \quad
    \begin{subfigure}[t]{0.3\columnwidth}
    \resizebox{1.1\columnwidth}{!}{
        \centering
        \begin{tikzpicture}
    \begin{axis}[
        xlabel={H multiplier},
        xmin=1, xmax=2,
        ymin=0, ymax=13,
        xtick={1, 1.1, 1.2, 1.3, 1.4, 1.5, 1.6, 1.7, 1.8, 1.9, 2},
        ytick={0, 2, 4, 6, 8, 10, 12},
        legend style={xshift=1.2cm},
        legend pos=north west,
        ymajorgrids=true,
        grid style=dashed,
    ]

    \addplot[
        color=blue,
        dashed,
        ]
        coordinates {
    (1, 0) (1.1, 4.9) (1.2, 6.5) (1.3, 7) (1.4, 8) (1.5, 8.6) (1.6, 9.6) (1.7, 9.6) (1.8, 9.6) (1.9, 10) (2, 10.1)       };

    \addplot[
        color=red,
        ]
        coordinates { 
    (1, 8) (1.1, 6) (1.2, 5) (1.3, 5) (1.4, 5) (1.5, 4) (1.6, 3) (1.7, 3) (1.8, 3) (1.9, 3) (2, 2.9) 
            };

            \addplot[
        color=green,
        ]
        coordinates { 
    (1, 5.1) (1.1,5) (1.2, 5) (1.3, 5) (1.4, 5) (1.5, 4.5) (1.6, 4.5) (1.7, 4.3) (1.8, 4.3) (1.9, 4.3) (2, 4.3) 
            };

            \addplot[
        ]
        coordinates { 
    (1, 4.8) (1.1, 4.8) (1.2, 4.8) (1.3, 4.8) (1.4, 4.8) (1.5, 4.8) (1.6, 4.8) (1.7, 4.8) (1.8, 4.8) (1.9, 4.3) (2, 4.3) 
            };
            
            \addplot[
        color=orange,
        ]
        coordinates {
    (1, 4) (1.1, 4) (1.2, 4) (1.3, 4) (1.4, 5) (1.5, 5) (1.6, 4) (1.7, 4) (1.8, 4) (1.9, 3) (2, 2) 
            };
            \addplot[
        color=purple,
        ]
        coordinates {
    (1, 4.8) (1.1, 4.8) (1.2, 4.8) (1.3, 4.8) (1.4, 4.8) (1.5, 4.8) (1.6, 4.4) (1.7, 4.4) (1.8, 4) (1.9, 3.9) (2, 2.6) 
            };
    \end{axis}
    \end{tikzpicture}
    }
        \subcaption{FJSSP}
    \end{subfigure}
    \quad
    \begin{subfigure}[t]{0.3\columnwidth}
    \resizebox{1.1\columnwidth}{!}{
        \centering
        \begin{tikzpicture}
    \begin{axis}[
        xlabel={H multiplier},
        xmin=1, xmax=2,
        ymin=0, ymax=500,
        xtick={1, 1.1, 1.2, 1.3, 1.4, 1.5, 1.6, 1.7, 1.8, 1.9, 2},
        ytick={0, 100, 200, 300, 400, 500},
        legend style={xshift=4cm},
        legend pos=north west,
        ymajorgrids=true,
        grid style=dashed,
    ]

    \addplot[
        dashed,
        color=blue,
        ]
        coordinates {
    (1, 0.0) (1.1, 479.3) (1.2, 493.2) (1.3, 493.2) (1.4, 493.2) (1.5, 493.2) (1.6, 493.2) (1.7, 493.2) (1.8, 493.2) (1.9, 493.2) (2, 493.2)        };

\addplot[
        color=red,
        ]
        coordinates {
    (1, 3.2) (1.1, 58.7) (1.2, 6.2) (1.3, 4.0) (1.4, 4.0) (1.5, 4.0) (1.6, 4.0) (1.7, 4.0) (1.8, 4.0) (1.9, 4.0) (2, 4.0) 
            };

            \addplot[
        color=green,
        ]
        coordinates {
    (1, 7.6) (1.1, 27.1) (1.2, 10.2) (1.3, 8.0) (1.4, 8.0) (1.5, 8.0) (1.6, 8.0) (1.7, 8.0) (1.8, 8.0) (1.9, 4.0) (2, 4.0) 
            };
                \addplot[
        ]
        coordinates {
    (1, 11.2) (1.1, 30.8) (1.2, 14.1) (1.3, 12.0) (1.4, 12.0) (1.5, 12.0) (1.6, 12.0) (1.7, 12.0) (1.8, 12.0) (1.9, 12.0) (2, 12.0) 
            };
    \addplot[
        color=orange,
        ]
        coordinates {
    (1, 7.8) (1.1, 8.3) (1.2, 2) (1.3, 2.0) (1.4, 2.0) (1.5, 2.0) (1.6, 2.0) (1.7, 2.0) (1.8, 2.0) (1.9, 2.0) (2, 2.0)
            };
            \addplot[
        color=purple,
        ]
        coordinates {
    (1, 7.6) (1.1, 3.3) (1.2, 2.0) (1.3, 2.0) (1.4, 2.0) (1.5, 2.0) (1.6, 2.0) (1.7, 2.0) (1.8, 2.0) (1.9, 2.0) (2, 2.0) 
            };
    
        \legend{$Régin$, $C$, $O$, $C \& O$, $Deg$, $R$}
        
    \end{axis}
    \end{tikzpicture}
    }
        \subcaption{StockingCost}
    \end{subfigure}
   \caption{Evolution of the average number of useless path computations for CHILD, FJSSP and StockingCost instances in function of the multiplier of $H$. The experimentation involves 4 landmarks.}
    \label{fig:averageNumberOfUselessArcInFunctionOfH}
\end{figure}
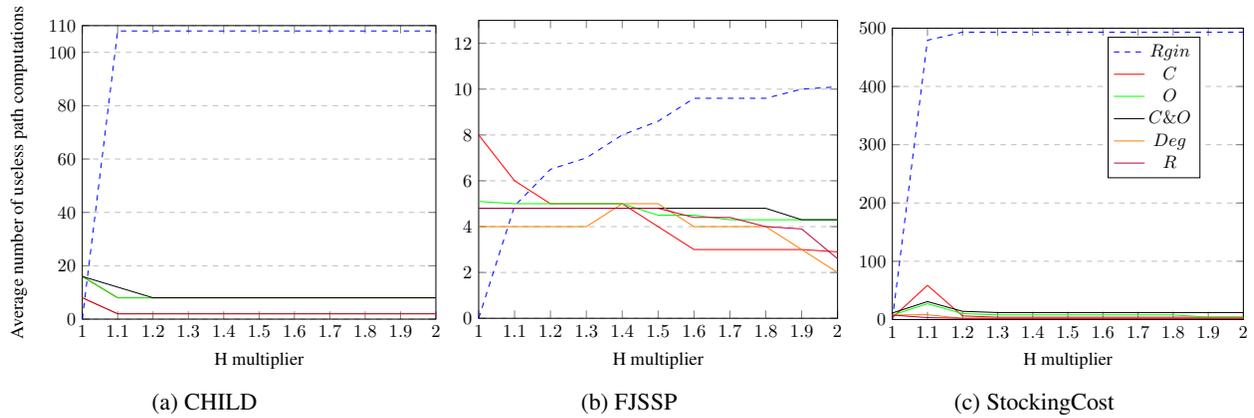

Figures~\ref{fig:averageNumberOfRemovedArcInFunctionOfH} and~\ref{fig:averageNumberOfUselessArcInFunctionOfH}  provide information on the relationship between H values and the number of useless path computations. The landmark approach performs very well as soon as the H value deviates a little from the optimal value, in other words, as soon as there is a little margin and therefore fewer inconsistent values. 

\subsubsection{TSP results}

\begin{figure}[h!]
    \centering

    \begin{tikzpicture}
    \begin{axis}[
        xlabel={Number of cities},
        ylabel={Time (ms)},
        xmin=0, xmax=1500,
        ymin=0, ymax=90000000000,
        xtick={0, 100, 300, 500, 700, 900, 1100, 1300, 1500},
        ytick={10000000000,20000000000,30000000000,40000000000,50000000000, 60000000000, 70000000000, 80000000000, 90000000000},
        legend pos=north west,
        ymajorgrids=true,
        grid style=dashed,
    ]

    \addplot[
        color=blue,
        mark=o,
        ]
        coordinates {
    (14, 2220530) (16, 950000) (17, 860060) (21, 1116660) (24, 930200)
(24, 1197780) (26, 1486690) (29, 1601580) (42, 2264490)
(42, 2446820)
(48, 3118910)
(48, 3321220)
(48, 3078120)
(51, 3532080)
(52, 3958960)
(58, 4979160)
(70, 7667390)
(76, 8617660)
(96, 15848440)
(99, 17797560)
(100, 19894100)
(100, 16807600)
(100, 17191910)
(100, 15929530)
(100, 17003020)
(100, 16843160)
(101, 16479730)
(105, 18383100)
(107, 16535900)
(120, 27059660)
(124, 27427500)
(127, 30530280)
(130, 32565560)
(136, 38467470)
(137, 41165780)
(144, 37680510)
(150, 57056550)
(150, 50954690)
(150, 51381680)
(152, 49643640)
(159, 63188440)
(175, 64339780)
(180, 46947610)
(195, 110060270)
(198, 123414070)
(200, 121176790)
(200, 119243860)
(202, 113262300)
(225, 179283770)
(225, 182400100)
(226, 169735440)
(229, 202433870)
(262, 262275090)
(264, 214641530)
(280, 323887980)
(299, 419320610)
(318, 427294250)
(400, 964638060)
(417, 1030552820)
(431, 1183442850)
(439, 1229169640)
(442, 1274221000)
(493, 1801032960)
(532, 2370052470)
(535, 2266427160)
(535, 1297533110)
(561, 2353925570)
(574, 3013194430)
(575, 2942079820)
(654, 4967022570)
(657, 4805718030)
(666, 4732261640)
(783, 17263562430)
(1002, 30067701280)
(1291, 42474652160)
(1304, 50133037030)
(1400, 49894810930)
(1432, 87534613800)    };

\addplot[
        color=red,
        mark=x,
        ]
        coordinates {
        (14, 1227000)
(16, 1008700)
(17, 715060)
(21, 889450)
(24, 1101340)
(24, 1517620)
(26, 1705810)
(29, 2209810)
(42, 2610940)
(42, 2845160)
(48, 4204640)
(48, 3798670)
(48, 2919540)
(51, 4258940)
(52, 3812570)
(58, 2288940)
(70, 2172670)
(76, 9414520)
(96, 6725130)
(99, 20475770)
(100, 12123190)
(100, 4645240)
(100, 10093940)
(100, 3992920)
(100, 4676540)
(100, 6886930)
(101, 17637780)
(105, 2763470)
(107, 11454610)
(120, 16538910)
(124, 5575740)
(127, 6056090)
(130, 4051070)
(136, 41555070)
(137, 29285090)
(144, 5014980)
(150, 8717280)
(150, 9496670)
(150, 14327130)
(152, 9410730)
(159, 24582630)
(175, 6530950)
(180, 12352790)
(195, 99808970)
(198, 46652570)
(200, 10623980)
(200, 17352510)
(202, 30276280)
(225, 22996050)
(225, 58084130)
(226, 21101450)
(229, 30861490)
(262, 18296620)
(264, 34477980)
(280, 404324570)
(299, 59041250)
(318, 26636700)
(400, 47837750)
(417, 80016300)
(431, 88090960)
(439, 43850150)
(442, 386383150)
(493, 111820180)
(532, 79688030)
(535, 81409860)
(535, 73577830)
(561, 87960690)
(574, 87853650)
(575, 90140700)
(654, 272354330)
(657, 174685950)
(666, 180177550)
(783, 423349600)
(1002, 527210030)
(1291, 703587660)
(1304, 904640900)
(1400, 760029830)
(1432, 1111893000)
        };
        \legend{Régin, Landmarks}
        
    \end{axis}
    \end{tikzpicture}
    
    \caption{Evolution of the time in relation to the size of TSP instances. The landmarks are selected with the degree method and 4 landmarks. The blue plot with circles is the time of the Régin algorithm and the red with crosses is the time of the landmarks algorithm.}
    \label{fig:graphTime}
\end{figure}
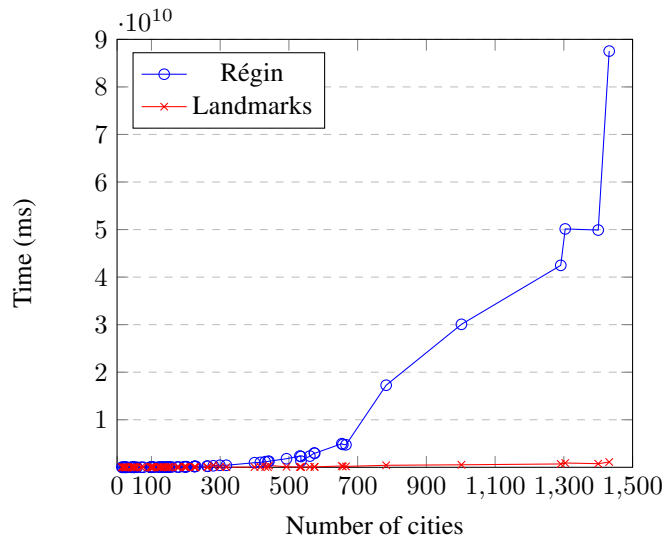

The improvement brought about by our approach for instances from the TSP problem are strong. This is mainly due to the relationship between the $H$ value given by the TSP value and the underlying costgcc constraint. In the case of the TSP, the optimal value of the min cost flow is lower than $H$, because the costgcc constraint only models a part of the problem and in fact represents a real relaxation. So even with an optimal $H$ value for TSP, there is a margin for the costgcc constraint.

To better appreciate the performance of the landmarks, Figure~\ref{fig:graphTime} shows the evolution of time in milliseconds as a function of the size of the TSP dataset instances. The blue plot with circles shows the time taken by the Régin's algorithm, while the red plot with crosses shows the time taken by the algorithm using the maximum degrees and 4 landmarks. Clearly, the use of landmarks is drastically faster than the Régin's algorithm. The larger the instance, the more useful landmarks become.

\begin{figure}[ht!]
    \centering
    \resizebox{0.6\columnwidth}{!}{

    \begin{tikzpicture}
    \begin{axis}[
        xlabel={Number of cities},
        ylabel={Time ratio (Régin/Landmarks)},
        xmin=0, xmax=1500,
        ymin=0, ymax=90,
        xtick={0, 100, 300, 500, 700, 900, 1100, 1300, 1500},
        ytick={0,20,40,60,80, 100},
        legend pos=north west,
        ymajorgrids=true,
        grid style=dashed,
    ]

    \addplot[
        color=blue,
        ]
        coordinates {
    (14,0.6) (16,0.4) (17,0.5) (21,0.4) (24,0.4) (24,0.6) (26,0.7) (29,0.6) (42,0.5) (42,0.6) (48,0.5) (48,0.6) (48,0.9) (51,0.8) (52,1.0) (58,1.6) (70,2.6) (76,0.7) (96,1.8) (99,0.8) (100,1.4) (100,3.1) (100,1.6) (100,3.9) (100,3.3) (100,2.4) (101,0.9) (105,6.4) (107,1.3) (120,1.3) (124,4.4) (127,4.8) (130,7.2) (136,1.0) (137,1.4) (144,7.7) (150,6.8) (150,5.9) (150,4.0) (152,5.9) (159,2.6) (175,9.1) (180,0.8) (195,1.1) (198,2.7) (200,12.0) (200,7.0) (202,4.3) (225,1.1) (225,3.3) (226,8.6) (229,6.8) (262,16.2) (264,6.7) (280,0.9) (299,9.4) (318,16.4) (400,21.8) (417,14.6) (431,14.8) (439,25.3) (442,3.5) (493,17.3) (532,33.0) (535,29.7) (535,17.6) (561,26.1) (574,23.8) (575,26.0) (654,20.1) (657,28.5) (666,27.9) (783,42.4) (1002,55.8) (1291,66.9) (1304,66.0) (1400,77.9) (1432,80.8)    };
        
    \end{axis}
    \end{tikzpicture}
    }
    \caption{Evolution of the speedup ratio in relation to the size of TSP instances. The landmarks are selected with the degree method and 4 landmarks.}
    \label{fig:graphRatio}

\end{figure}
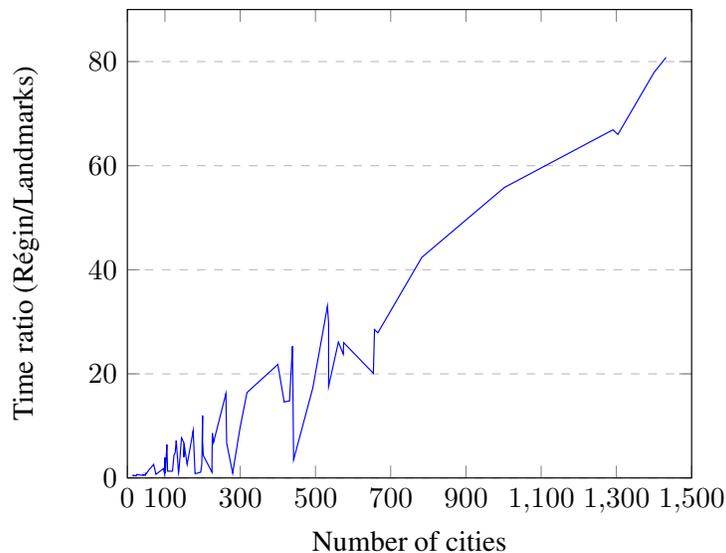

Figure~\ref{fig:graphRatio} shows the evolution of the speed-up ratio (Régin time/Landmarks time) on the instances of the TSP dataset. The landmark selection algorithm is based on maximum degrees with 4 landmarks. We can also see in this graph that the more data there is, the higher the gain factor. As mentioned above, this can be explained by the fact that in a large structure, the landmarks contain a lot of information compared with a smaller structure. 
In these experiments, note that if we omit the assigned variables there is only one strongly connected component in the value network. Overall, we find that our algorithm significantly speeds up the previous approach, up to about 80 times faster for large problems.

\newpage
\section{Conclusion}
\label{sec:Conclusion}

This paper proposes an efficient implementation of the arc consistency algorithm of the cardinality constraint with costs. 
This constraint is present in many industrial problems and the establishment of the arc consistency is often too slow to be used in practice, as it is based on finding the shortest paths from the assigned values. We introduce a new method that uses upper bounds on shortest paths based on triangular inequalities and landmarks. 
This approach avoids the computation of many shortest paths and improves the computation time of the arc consistency filtering algorithm.
The larger the graph and the larger the margins, the greater the improvement will be.
In addition, we have introduced a sufficient condition, which is quick to compute, for a costgcc constraint to be arc consistent.

\section*{Acknowledgments}
This work has been supported by the French government, through the 3IA Côte d’Azur Investments in the Future
project managed by the National Research Agency (ANR)
with the reference number ANR-19-P3IA-0002.

\bibliographystyle{unsrt}  
\bibliography{biblio}

\end{document}